\documentclass[letter, 10pt, conference]{ieeeconf}
\usepackage{times}
\def\IEEEPARstart#1#2{#1#2}
\long\def\comment#1\endcomment{}

\usepackage{multicol}

  \usepackage{amsmath,amssymb,amsfonts,bm}
  \def\tymes{{\times}}
  \usepackage[
    free-standing-units = true,
    per-mode = single-symbol,
    product-units = power,
    product-symbol = \tymes,
    range-units = single,
    list-units = single,
    list-final-separator = {, and },
    uncertainty-mode = separate,
    bracket-ambiguous-numbers = false
  ]{siunitx}
  \usepackage{graphicx}
  \usepackage{subcaption}

  \usepackage{xcolor}
  \usepackage[most]{tcolorbox}
  \usepackage{booktabs}
  \usepackage{balance}
  \usepackage{blindtext}
  \newif\ifhyperref{}\hyperreftrue{}
  \ifhyperref{}
    \usepackage[hidelinks,bookmarksopen=true,bookmarks=true]{hyperref}
    \usepackage[all]{hypcap}
  \else
    \DeclareRobustCommand{\href}[2]{#1}
    
  \fi
  \usepackage[nameinlink,capitalise]{cleveref}
  \usepackage{adjustbox}
  \usepackage{multirow}
  \usepackage{threeparttable}

  \usepackage{tikz}
  \usetikzlibrary{positioning, shapes.geometric, arrows.meta, chains, matrix, decorations.pathreplacing}

  \usepackage{pgfplots}
  \DeclareUnicodeCharacter{2212}{−}
  \usepgfplotslibrary{groupplots,dateplot}
  \usetikzlibrary{patterns,shapes.arrows}
  \usetikzlibrary{fit,shapes,calc,positioning}
  \pgfplotsset{compat=newest}

\everymath=\expandafter{\the\everymath\displaystyle}
\IfFileExists{scrextend.sty}{
  \usepackage[fontsize=10.000000pt]{scrextend}
}{
  \renewcommand{\normalsize}{\fontsize{10.000000}{12.000000}\selectfont}
  \normalsize
}
\usepackage{siunitx}
\DeclareSIUnit\cells{cells}

\ifdefined\pdftexversion\else  % non-pdftex case.
  \usepackage{fontspec}
\fi
\makeatletter\@ifpackageloaded{underscore}{}{\usepackage[strings]{underscore}}\makeatother

\def\pct#1{\qty{#1}{\percent}}

\DeclareRobustCommand{\eat}[1]{}

\def\tymes{{\times}}

\def\eg{e.g.}
\def\ie{i.e.}

\DeclareRobustCommand{\[}{\begin{equation}}
\DeclareRobustCommand{\]}{\end{equation}}
\def\eqnend#1{}

\def\constant#1{\ifmmode\text{`\MakeLowercase{#1}'}\else`\MakeLowercase{#1}'\fi}

\DeclareSIUnit\cells{cells}

\definecolor{changecolor}{RGB}{50,0,225} 
\DeclareMathOperator*{\argmax}{arg\,max}
\DeclareMathOperator*{\argmin}{arg\,min}
  \definecolor{mplC0}{HTML}{1F77B4}
\definecolor{mplC1}{HTML}{FF7F0E}
\definecolor{mplC2}{HTML}{2CA02C}
\definecolor{mplC3}{HTML}{D62728}
\definecolor{mplC4}{HTML}{9467BD}
\definecolor{mplC5}{HTML}{8C564B}
\definecolor{mplC6}{HTML}{E377C2}
\definecolor{mplC7}{HTML}{7F7F7F}
\definecolor{mplC8}{HTML}{BCBD22}
\definecolor{mplC9}{HTML}{17BECF}
\definecolor{mplaliceblue}{HTML}{F0F8FF}
\definecolor{mplantiquewhite}{HTML}{FAEBD7}
\definecolor{mplaqua}{HTML}{00FFFF}
\definecolor{mplaquamarine}{HTML}{7FFFD4}
\definecolor{mplazure}{HTML}{F0FFFF}
\definecolor{mplbeige}{HTML}{F5F5DC}
\definecolor{mplbisque}{HTML}{FFE4C4}
\definecolor{mplblack}{HTML}{000000}
\definecolor{mplblanchedalmond}{HTML}{FFEBCD}
\definecolor{mplblue}{HTML}{0000FF}
\definecolor{mplblueviolet}{HTML}{8A2BE2}
\definecolor{mplbrown}{HTML}{A52A2A}
\definecolor{mplburlywood}{HTML}{DEB887}
\definecolor{mplcadetblue}{HTML}{5F9EA0}
\definecolor{mplchartreuse}{HTML}{7FFF00}
\definecolor{mplchocolate}{HTML}{D2691E}
\definecolor{mplcoral}{HTML}{FF7F50}
\definecolor{mplcornflowerblue}{HTML}{6495ED}
\definecolor{mplcornsilk}{HTML}{FFF8DC}
\definecolor{mplcrimson}{HTML}{DC143C}
\definecolor{mplcyan}{HTML}{00FFFF}
\definecolor{mpldarkblue}{HTML}{00008B}
\definecolor{mpldarkcyan}{HTML}{008B8B}
\definecolor{mpldarkgoldenrod}{HTML}{B8860B}
\definecolor{mpldarkgray}{HTML}{A9A9A9}
\definecolor{mpldarkgreen}{HTML}{006400}
\definecolor{mpldarkgrey}{HTML}{A9A9A9}
\definecolor{mpldarkkhaki}{HTML}{BDB76B}
\definecolor{mpldarkmagenta}{HTML}{8B008B}
\definecolor{mpldarkolivegreen}{HTML}{556B2F}
\definecolor{mpldarkorange}{HTML}{FF8C00}
\definecolor{mpldarkorchid}{HTML}{9932CC}
\definecolor{mpldarkred}{HTML}{8B0000}
\definecolor{mpldarksalmon}{HTML}{E9967A}
\definecolor{mpldarkseagreen}{HTML}{8FBC8F}
\definecolor{mpldarkslateblue}{HTML}{483D8B}
\definecolor{mpldarkslategray}{HTML}{2F4F4F}
\definecolor{mpldarkslategrey}{HTML}{2F4F4F}
\definecolor{mpldarkturquoise}{HTML}{00CED1}
\definecolor{mpldarkviolet}{HTML}{9400D3}
\definecolor{mpldeeppink}{HTML}{FF1493}
\definecolor{mpldeepskyblue}{HTML}{00BFFF}
\definecolor{mpldimgray}{HTML}{696969}
\definecolor{mpldimgrey}{HTML}{696969}
\definecolor{mpldodgerblue}{HTML}{1E90FF}
\definecolor{mplfirebrick}{HTML}{B22222}
\definecolor{mplfloralwhite}{HTML}{FFFAF0}
\definecolor{mplforestgreen}{HTML}{228B22}
\definecolor{mplfuchsia}{HTML}{FF00FF}
\definecolor{mplgainsboro}{HTML}{DCDCDC}
\definecolor{mplghostwhite}{HTML}{F8F8FF}
\definecolor{mplgold}{HTML}{FFD700}
\definecolor{mplgoldenrod}{HTML}{DAA520}
\definecolor{mplgray}{HTML}{808080}
\definecolor{mplgreen}{HTML}{008000}
\definecolor{mplgreenyellow}{HTML}{ADFF2F}
\definecolor{mplgrey}{HTML}{808080}
\definecolor{mplhoneydew}{HTML}{F0FFF0}
\definecolor{mplhotpink}{HTML}{FF69B4}
\definecolor{mplindianred}{HTML}{CD5C5C}
\definecolor{mplindigo}{HTML}{4B0082}
\definecolor{mplivory}{HTML}{FFFFF0}
\definecolor{mplkhaki}{HTML}{F0E68C}
\definecolor{mpllavender}{HTML}{E6E6FA}
\definecolor{mpllavenderblush}{HTML}{FFF0F5}
\definecolor{mpllawngreen}{HTML}{7CFC00}
\definecolor{mpllemonchiffon}{HTML}{FFFACD}
\definecolor{mpllightblue}{HTML}{ADD8E6}
\definecolor{mpllightcoral}{HTML}{F08080}
\definecolor{mpllightcyan}{HTML}{E0FFFF}
\definecolor{mpllightgoldenrodyellow}{HTML}{FAFAD2}
\definecolor{mpllightgray}{HTML}{D3D3D3}
\definecolor{mpllightgreen}{HTML}{90EE90}
\definecolor{mpllightgrey}{HTML}{D3D3D3}
\definecolor{mpllightpink}{HTML}{FFB6C1}
\definecolor{mpllightsalmon}{HTML}{FFA07A}
\definecolor{mpllightseagreen}{HTML}{20B2AA}
\definecolor{mpllightskyblue}{HTML}{87CEFA}
\definecolor{mpllightslategray}{HTML}{778899}
\definecolor{mpllightslategrey}{HTML}{778899}
\definecolor{mpllightsteelblue}{HTML}{B0C4DE}
\definecolor{mpllightyellow}{HTML}{FFFFE0}
\definecolor{mpllime}{HTML}{00FF00}
\definecolor{mpllimegreen}{HTML}{32CD32}
\definecolor{mpllinen}{HTML}{FAF0E6}
\definecolor{mplmagenta}{HTML}{FF00FF}
\definecolor{mplmaroon}{HTML}{800000}
\definecolor{mplmediumaquamarine}{HTML}{66CDAA}
\definecolor{mplmediumblue}{HTML}{0000CD}
\definecolor{mplmediumorchid}{HTML}{BA55D3}
\definecolor{mplmediumpurple}{HTML}{9370DB}
\definecolor{mplmediumseagreen}{HTML}{3CB371}
\definecolor{mplmediumslateblue}{HTML}{7B68EE}
\definecolor{mplmediumspringgreen}{HTML}{00FA9A}
\definecolor{mplmediumturquoise}{HTML}{48D1CC}
\definecolor{mplmediumvioletred}{HTML}{C71585}
\definecolor{mplmidnightblue}{HTML}{191970}
\definecolor{mplmintcream}{HTML}{F5FFFA}
\definecolor{mplmistyrose}{HTML}{FFE4E1}
\definecolor{mplmoccasin}{HTML}{FFE4B5}
\definecolor{mplnavajowhite}{HTML}{FFDEAD}
\definecolor{mplnavy}{HTML}{000080}
\definecolor{mploldlace}{HTML}{FDF5E6}
\definecolor{mplolive}{HTML}{808000}
\definecolor{mplolivedrab}{HTML}{6B8E23}
\definecolor{mplorange}{HTML}{FFA500}
\definecolor{mplorangered}{HTML}{FF4500}
\definecolor{mplorchid}{HTML}{DA70D6}
\definecolor{mplpalegoldenrod}{HTML}{EEE8AA}
\definecolor{mplpalegreen}{HTML}{98FB98}
\definecolor{mplpaleturquoise}{HTML}{AFEEEE}
\definecolor{mplpalevioletred}{HTML}{DB7093}
\definecolor{mplpapayawhip}{HTML}{FFEFD5}
\definecolor{mplpeachpuff}{HTML}{FFDAB9}
\definecolor{mplperu}{HTML}{CD853F}
\definecolor{mplpink}{HTML}{FFC0CB}
\definecolor{mplplum}{HTML}{DDA0DD}
\definecolor{mplpowderblue}{HTML}{B0E0E6}
\definecolor{mplpurple}{HTML}{800080}
\definecolor{mplrebeccapurple}{HTML}{663399}
\definecolor{mplred}{HTML}{FF0000}
\definecolor{mplrosybrown}{HTML}{BC8F8F}
\definecolor{mplroyalblue}{HTML}{4169E1}
\definecolor{mplsaddlebrown}{HTML}{8B4513}
\definecolor{mplsalmon}{HTML}{FA8072}
\definecolor{mplsandybrown}{HTML}{F4A460}
\definecolor{mplseagreen}{HTML}{2E8B57}
\definecolor{mplseashell}{HTML}{FFF5EE}
\definecolor{mplsienna}{HTML}{A0522D}
\definecolor{mplsilver}{HTML}{C0C0C0}
\definecolor{mplskyblue}{HTML}{87CEEB}
\definecolor{mplslateblue}{HTML}{6A5ACD}
\definecolor{mplslategray}{HTML}{708090}
\definecolor{mplslategrey}{HTML}{708090}
\definecolor{mplsnow}{HTML}{FFFAFA}
\definecolor{mplspringgreen}{HTML}{00FF7F}
\definecolor{mplsteelblue}{HTML}{4682B4}
\definecolor{mpltan}{HTML}{D2B48C}
\definecolor{mplteal}{HTML}{008080}
\definecolor{mplthistle}{HTML}{D8BFD8}
\definecolor{mpltomato}{HTML}{FF6347}
\definecolor{mplturquoise}{HTML}{40E0D0}
\definecolor{mplviolet}{HTML}{EE82EE}
\definecolor{mplwheat}{HTML}{F5DEB3}
\definecolor{mplwhite}{HTML}{FFFFFF}
\definecolor{mplwhitesmoke}{HTML}{F5F5F5}
\definecolor{mplyellow}{HTML}{FFFF00}
\definecolor{mplyellowgreen}{HTML}{9ACD32}

\begin{document}

  % front matter

  %\def\t{Floor Plans for Predicting Occupancy using Sensor History: A Language Model Approach}
  %\def\t{Towards Online Predictive Exploration and Path Planning: A Synthetic Dataset}
  %\def\t{From Sensors to Sensing: Predicting Unexplored Indoor Environments from Sensor Data}
  %\def\t{Predicting Unexplored Indoor Environments from Sensor Data}
  %\def\t{Predicting Partially Observed Indoor Environments from Sensor Data by Learning from Floor Plans}
  %\def\t{MATE: Multi-Agent-in-Time Autonomous Exploration}
  %\def\t{Autonomous Exploration is Multiple Personality Disorder}
  %\def\t{Robots Contain Multitudes: Multi-Agent-in-Time Autonomous Exploration}
  %\def\t{\textit{The Road Less Traveled} -- Autonomous Exploration as Planning over Path Densities}
  %\def\t{Frontier Size Matters: Information Gain Is Not All You Need}
  \def\t{Information Gain Is Not All You Need}
  \def\a{\renewcommand\thefootnote{*}\footnotetext{Equal contribution.}Ludvig Ericson\footnotemark{}, José Pedro\footnotemark{}, Patric Jensfelt}

  %A Vector-based Generative Model of Partial Floor Plans}
  \title{\t}

  %\author{\a%
  %\thanks{[Placeholder] Paper accepted for the 99th Earth Conference on Papers (ECP 3000). All authors are with the Division of Robotics, Perception and Learning at KTH Royal Institute of Technology, Stockholm, SE-10044, Sweden. This work was supported by the the VR grant XPLORE3D. For e-mail correspondence, contact {\tt\small{ludv@kth.se}}. \newline 123-1-1234-1234-1/12/\$12.34 \textcopyright{} 1234 ABCD.}
  %}

  \ifhyperref
    \hypersetup{%
      pdftitle=\t,
      pdfauthor=\a,
      pdfcreator=,
      bookmarksopenlevel=2
    }
  \fi

  \newif\ifabstract{}

  %\abstractfalse{}
  \abstracttrue{}

  %\def\funkyfiggraphics{
  %  \refstepcounter{figure}\label{fig:roomplan}%
  %  \centering{%
  %  %\includegraphics[trim=0 5.5cm 0 5.5cm,clip,width=0.99\linewidth]{roomplan/floorist_flow_serif.pdf}%
  %  %\includegraphics[trim=5.5cm 5.2cm 5.5cm 5.2cm,clip,width=0.99\linewidth]{roomplan/floorist_flow_2part.pdf}
  %  \includegraphics[trim=68mm 68mm 70mm 70mm,clip,width=0.90\linewidth]{roomplan/floorist_flow_2part2.pdf}%
  %  }}
  %%Illustration of going from sensors to sensing. A sensor, in this case a smartphone with a LIDAR using RoomPlan, an off-the-shelf floor plan reconstruction solution, is used to create a parametric 3D model of an indoor environment. A virtual robot is simulated to construct an occupancy grid, which is then input to our predictive model to ``sense'' what is beyond the observed space.}
  %\def\funkyfigcaptiontext{%
  %  Top: An illustration of the real-world pipeline. A sensor, in this case a
  %  smartphone with a LIDAR using RoomPlan, an off-the-shelf floor plan mapping
  %  solution, is used to reconstruct a parametric 3D model of an indoor
  %  environment. A virtual robot is simulated in the reconstructed environment
  %  and produces an occupancy grid, which is then used by our model Floorist to
  %  ``sense'' what is beyond the observed space. Bottom: An illustration of the
  %  synthesizing pipeline used for training. Starting with a database of floor
  %  plans, runs of a virtual robot are recorded as a set of occupancy grid
  %  snapshots. The unseen walls of the floor plan are the targets of a
  %  statistical model given the occupancy grid.}%

  \author{
    %\thanks{Manuscript received: January, 23, 2024; Revised April, 8, 2024; Accepted May, 12, 2024.
    %This paper was recommended for publication by Hanna Kurniawati upon evaluation of the Associate Editor and Reviewers' comments.}%Use only for final RAL version
    %\thanks{Digital Object Identifier (DOI): see top of this page.}%
    Ludvig Ericson\textsuperscript{*,1}, José Pedro\textsuperscript{*,1,2}, Patric Jensfelt\textsuperscript{1}
  }

  \maketitle

  \ifabstract{}
  \begin{abstract}
    Autonomous exploration in mobile robotics often involves a trade-off between two objectives: maximizing environmental coverage and minimizing the total path length. In the widely used information gain paradigm, exploration is guided by the expected value of observations. While this approach is effective under \emph{budget-constrained} settings---where only a limited number of observations can be made---it fails to align with \emph{quality-constrained} scenarios, in which the robot must fully explore the environment to a desired level of certainty or quality. In such cases, total information gain is effectively fixed, and maximizing it per step can lead to inefficient, greedy behavior and unnecessary backtracking.
    This paper argues that information gain should not serve as an optimization objective in quality-constrained exploration. Instead, it should be used to filter viable candidate actions. We propose a novel heuristic, \emph{distance advantage}, which selects candidate frontiers based on a trade-off between proximity to the robot and remoteness from other frontiers. This heuristic aims to reduce future detours by prioritizing exploration of isolated regions before the robot's opportunity to visit them efficiently has passed.
    We evaluate our method in simulated environments against classical frontier-based exploration and gain-maximizing approaches. Results show that distance advantage significantly reduces total path length across a variety of environments, both with and without access to prior map predictions. Our findings challenge the assumption that more accurate gain estimation improves performance and offer a more suitable alternative for the quality-constrained exploration paradigm.
  \end{abstract}
  \fi

\IEEEpeerreviewmaketitle

  {\def\thefootnote{}%
    \footnotetext{\textsuperscript{*} Equal contribution.
    
    \textsuperscript{1} Authors are with the Division of Robotics, Perception and Learning at KTH Royal Institute of Technology, Stockholm, SE-10044, Sweden.
    
    \textsuperscript{2} Author is with Ericsson Research, Stockholm, Sweden.
    
    This work was supported by the Swedish Research Council, and the Wallenberg AI, Autonomous Systems and Software Program (WASP) funded by the Knut and Alice Wallenberg Foundation.
    For e-mail correspondence, contact {\tt\footnotesize{ludv@kth.se}}.
    Experiments were run on the Berzelius
cluster provided by NSC at Linköping University, Sweden.}
  }%

\section{Introduction}

%Autonomous exploration in mobile robotics is the problem of efficiently collecting observations from an unknown environment. It is fundamental in a wide variety of applications ranging from search and rescue \shite{} to industrial inspection \shite{} and active reconstruction.

Autonomous exploration is a fundamental problem in mobile robotics, found in a wide variety of applications ranging from search and rescue \cite{calisi-autonomous-2007, colas-3d-2013} to industrial inspection \cite{omari-visual-2014}. Its definition varies between applications, and we identify two main reoccurring variations: \emph{budget-constrained exploration}, and \emph{quality-constrained exploration}.
In budget-constrained exploration, it is assumed that the robot's budget is limited, e.g., by its battery, a deadline, or a camera roll. The budget is insufficient to fully cover the environment and the optimal set of views must be selected.
In quality-constrained exploration, the constraint is flipped and it is assumed that the robot can cover its environment, and instead must collect views such that the whole map is of some minimum level of quality, i.e., the remaining \emph{gain} falls below a given threshold.

Crucially, no part of the environment is left unexplored in quality-constrained exploration, implying that the total gain is essentially constant.
The rich body of exploration methods that maximize gain must therefore be said to implicitly assume a budget-constrained scenario, because to maximize gain per sensor scan is to assume that some gain will be left unexplored, and that total gain is not constant.

\begin{figure}[tb]
  \centering%
  \iffalse
  \begin{adjustbox}{clip,trim=2mm 3mm 2mm 2mm,width=\linewidth}%
    \input{ig_lambda.pgf}%
  \end{adjustbox}%
  \fi%
  \includegraphics[clip,trim=2mm 3mm 2mm 2mm,width=\linewidth]{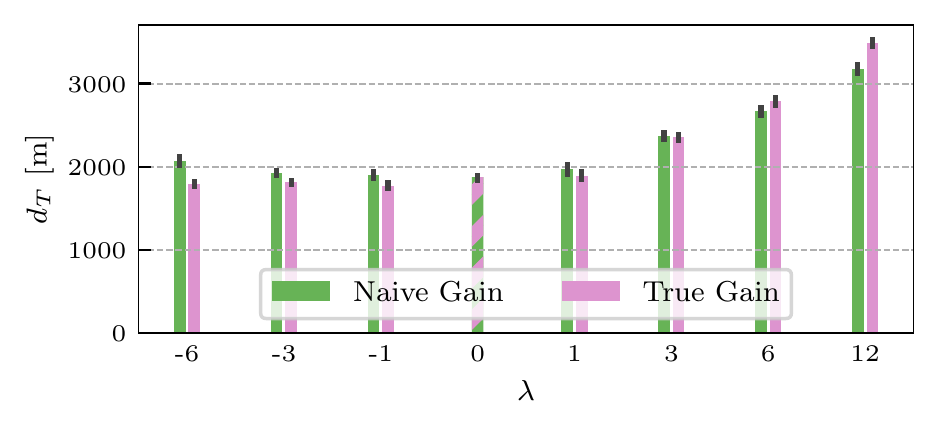}%
  \caption{
  Distance at completion $d_T$ for a selection of gain affinities $\lambda$, where a higher $\lambda$ means stronger preference for gain and a lesser concern with the length of the path to acquire it. Naive gain refers to the assumption that unknown space is occlusion-free, i.e., yields maximal gain; in true gain, the real would-be sensor scan is used for gain computation.
  %Higher affinity leads to longer paths, and the effect is stronger with access to the actual gain.
  Tellingly, negative affinities, \ie, \textit{minimizing} gain, results in a lower $d_T$ than maximization, and no choice is substantially better than nearest frontier, \ie, $\lambda=0$.
  \iffalse
  Gain maximization methods in autonomous exploration attempt to establish a trade-off between the gain $G(f)$ obtained by following a path to a frontier $f$, and the cost $d(f)$ of that path.
  The cost function maximized is usually of the form ${\lambda \log G(f) - d(f)}$, where $\lambda$ is the gain affinity that regulates the preference of gain over path cost.
  The path length at exploration completion was evaluated for different values for the affinity $\lambda$ and considering a typical gain estimator, naive, and an oracle.
  These results show that when aiming to completely explore the environment, higher affinity consistently leads to longer paths, an effect that is worsened with access to the perfect predictions of the gain.
  Tellingly, negative affinity, \ie, gain \textit{minimization}, achieves slightly shorter paths than maximization, if there.
  \fi
  }%
  \label{fig:ig-lambda}
\end{figure}

%€It has been observed that maximizing gain in a quality-constrained paradigm, by prioritizing information gain over path cost \cite{obwald-speeding-up-2016} or accurately predicting information gain \cite{ericson-understanding-2021}, can counterintuitively lead to longer paths.
The two paradigms are sometimes mixed up, such as by using a method suited for budget-constrained exploration but evaluating its quality-constrained properties, e.g., evaluating gain maximization by path length.
In many cases, authors report counterintuitive results such as longer total path lengths and greedy behavior with improved gain estimation, owing to unnecessary backtracking as exploration draws to a finish \cite{ericson-understanding-2021, luperto-estimating-2024, obwald-speeding-up-2016}.
\Cref{fig:ig-lambda} shows that path length increases with stronger affinity to maximize gain, and that gain should in fact be \emph{minimized}, if at all considered.
This incongruence ultimately stems from the mix-up of the budget-constrained and quality-constrained paradigms: in budget-constrained exploration, obtaining gain is the priority, so greediness is, in fact, the desired outcome.
Relatively little prior work can therefore be said to be targeted at the quality-constrained paradigm directly, which is the aim of this paper.

While some backtracking is inevitable, there can also be unnecessary backtracking.
\cite{obwald-speeding-up-2016}~show that unnecessary backtracking is incurred if and only if a given region is not yet explored by the \emph{last} time that the optimal plan passes it.
In other words, the robot typically has multiple opportunities to explore a given region, and unnecessary backtracking is guaranteed only once it has missed its last chance.

The optimal plan is, of course, not available to the robot, and it cannot know when it has its last opportunity to explore a region. A cue that this might be the case, however, is that the robot is closer to that region now than it is expected to be later; it has a rare opportunity to explore the region with low path cost. We dub this heuristic \emph{distance advantage}.

%Isolated frontiers are less likely to be revisited; therefore, visiting these frontiers first should minimize backtracking.

The phenomenon where gain maximization causes worse performance was first identified in works aiming to improve exploration by leveraging additional knowledge~\cite{obwald-speeding-up-2016,ericson-beyond-2024}; consequently, in this paper, we propose a method for autonomous exploration that does not maximize gain, and instead centers on reducing backtracking by leveraging prior knowledge through our distance advantage heuristic.
%{\remark{[→ related work?]}}
%Some works \cite{obwald-speeding-up-2016} have considered the complete coverage paradigm for autonomous exploration.
%However, they assume some kind of prior knowledge about the \textit{whole} environment, which is often not available in autonomous exploration.
%The only kind of prior knowledge that can always be expected is common sense, made available through statistical models \shite{}, but these only provide local predictions about the unknown environment.

The paper is outlined as follows: first, related work is analyzed through the lens of budget contra quality-constrained exploration in~\cref{sec:related}, the quality-constrained exploration problem is then formally defined in~\cref{sec:problem}. Next, our proposed method is presented in~\cref{sec:da}, the experimental setup and results in~\cref{sec:experimental-setup,sec:results,sec:predictions}, and finally, our conclusions and limitations are presented in~\cref{sec:limitations,sec:conclusion}. To summarize, our main contributions are:
\begin{itemize}
    \item the relationship between autonomous exploration and gain maximization is disentangled once and for all, explaining and justifying counterintuitive findings reported in existing literature;
    \item a novel method for autonomous exploration, distance advantage, is proposed, which directly aims to minimize future potential backtracking; and
    \item its performance is evaluated in simulation, producing significantly shorter paths in the quality-constrained paradigm than nearest frontier~\cite{yamauchi-frontier-based-1997} or gain maximization~\cite{gonzalez-banos-navigation-2002}.
\end{itemize}

The reader is encouraged to review the
supplementary video material for this paper, as it offers insights that are difficult to convey with text and images. 
This material, along with an implementation of the proposed method, is available at \underline{\href{https://lericson.se/da/}{lericson.se/da}}.

\section{Related Work}\label{sec:related}

Autonomous exploration was initially proposed in the context of active perception \cite{bajcsy-active-1988}, with the term being coined by Whaite and Ferry \cite{whaite-autonomous-1997}.
In many tasks, a single measurement is insufficient, either due to the nature of the sensor or due to uncertainty.
Therefore, it is necessary to plan where to place the sensor in order to collect measurements that most reduce uncertainty.
Since the focus of active perception was on object reconstruction in a small workspace, the cost of the path necessary to move the sensor was not relevant.
Therefore, these first works were direct extensions of the next-best-view \cite{connolly-determination-1985} methods from vision to robotics, by maximizing the predicted gain of the next measurement \cite{bajcsy-active-1988, maver-occlusions-1993, pito-sensor-based-1996, whaite-autonomous-1997}.

The subtle confusion between budget-constrained exploration and quality-constrained exploration was already present in the active perception community.
Whaite and Ferry \cite{whaite-autonomous-1997} defined the goal of autonomous exploration as that of determining representations of acceptable fidelity, i.e., quality-constrained exploration.
However, they proposed to approach autonomous exploration through the lens of gain maximization, implicitly adopting the budget-constrained paradigm.

Carrying over to mobile robotics, where the cost of moving the robot cannot be neglected, the focus became to maximize gain with respect to the distance traveled \cite{gonzalez-banos-navigation-2002, yamauchi-frontier-based-1997}, remaining in the implicit budget-constrained paradigm.
Finding the best trade-off between gathering information and moving efficiently has proven to be a challenging problem, leading to extensive research and the very term `autonomous exploration' being appropriated by the mobile robotics community.

Yamauchi \cite{yamauchi-frontier-based-1997} argued that to efficiently observe the environment, the robot should plan to visit states that are predicted to have high gain while minimizing the path cost.
The concept of frontiers, the border between known and unknown space, was introduced and it was proposed that exploration be done by navigating to the nearest frontier.
Directly extending the next-best-view approaches, \cite{gonzalez-banos-navigation-2002} proposed to explicitly optimize for measurement gain, weighted inversely to the distance necessary to collect it.
Already when introducing gain maximization for mobile robotics, \cite{gonzalez-banos-navigation-2002} observed that prioritizing gain led to fast short-term exploration, but ultimately made completing exploration take longer.

A wide body of literature exists building on \cite{gonzalez-banos-navigation-2002, yamauchi-frontier-based-1997}, extending it to more complex scenarios and improving upon its assumptions.
RH-NBV \cite{bircher-receding-2016} evaluates the objective along a \textit{path}, instead of a single step.
While searching over paths scales combinatorially, when compared to single decisions, this can be dealt with through sampling-based planning.
AEP \cite{selin-efficient-2019} combines frontiers and RH-NBV to mitigate the effect of greediness due to gain maximization in the global path, while preserving the efficient local coverage performance of RH-NBV.
Some works \cite{kulich-2011-distance,zhou-fuel-2021}, most notably FUEL \cite{zhou-fuel-2021}, attempt to plan optimal tours for a Traveling Salesman Problem (TSP) that visits every frontier cluster. 
The gain of each frontier cluster can also be locally optimized in a refinement step \cite{zhou-fuel-2021}.
RH-NBV and AEP require extensive gain estimation, which can take up to 95\% of planning time \cite{schmid-efficient-2020}, while FUEL computes solutions to a TSP, which can be computationally demanding.
Returning to single-step planning, UFOExplorer \cite{duberg-ufoexplorer-2022} shows that determining frontiers to be states with a minimum amount of gain, the nearest frontier exploration strategy produces shorter paths than RH-NBV, AEP and FUEL, while being computationally cheaper.
ECHO \cite{jiajie-echo-2023} combines \cite{duberg-ufoexplorer-2022} and \cite{zhou-fuel-2021}, choosing the nearest gain-having frontier and then optimizing the viewpoint to improve gain.
More works extend these ideas, by improving upon the sampling-based planner and the path cost function \cite{lindqvist-explorationrrt-2021, schmid-efficient-2020,witting-history-2018}, or considering improved estimates of gain through learning-based methods \cite{deng-robotic-2020, schmid-fast-2022, shrestha-learned-2019, tao-seer-2023}, among others.

Importantly, most works aim to efficiently complete coverage of unknown environments and therefore evaluate quality-constrained exploration metrics, such as time or path length.
However, they are implicitly performing budget-constrained exploration since they maximize gain.
Several counterintuitive results have been reported, such as that maximizing gain ultimately leads to longer coverage paths \cite{gonzalez-banos-navigation-2002, obwald-speeding-up-2016}, or that improved estimates of gain lead to longer paths \cite{ericson-understanding-2021}.

% A source of difficulty in interpreting these results stems from the optimal exploration path being intractable to determine, even if attempted offline and with full knowledge of the environment.
% \cite{li-searching-2012} has attempted to provide a bound on optimal performance and estimate the competitive ratio of existing strategies, but this evaluation has not been widely adopted by the community and it is typically not known how much room for improvement there is in exploration.

Few works have explicitly addressed the mismatch between quality-constrained exploration goals and the budget-constrained optimization objectives.
Li \textit{et al} \cite{li-searching-2012} attempted to determine the optimal exploration path offline by using A$^*$ with an admissible heuristic, estimated from a complete map of the environment.
The determined path is used to estimate the competitive ratio of other exploration strategies, showing that pure gain maximization is far from optimal.
When an abstract topological map of the environment is available, a high-level global exploration path can be determined by solving a TSP \cite{obwald-speeding-up-2016}.
The solution to the TSP can be used to determine the optimal order in which to visit frontiers, potentially using information gain to locally prioritize frontiers, similar to \cite{selin-efficient-2019}.
Ultimately, it is shown that using information gain leads to longer paths, and that strictly prioritizing the order in which frontiers are visited based on the TSP solution is better, since exploration is finished only when there are no more frontiers.
Since more information about the environment does not lead to shorter paths for information gain, \cite{ericson-understanding-2021} proposed a heuristic that prioritizes visiting frontiers that observe elements of the environment that cannot be observed together with other elements.
This is shown to lead to shorter paths than gain maximization and to improve with access to more information about the environment.

Combining information gain with quality-driven exploration is possible, and, in fact, paramount to tasks like active SLAM.
There, exploration must compete with active localization: the goal is to produce a sufficiently high-quality map, but that requires maintaining an accurate state estimate.
Works like \cite{stachniss-information-2005, yichen-exploration-2022} address this scenario, where the role of information gain is not to drive exploration, but instead as a form of exploitation for minimizing state uncertainty.

\section{Problem Statement}\label{sec:problem}

A plan $\pi$ is a sequence of states, i.e., $\pi=(s_0,\ldots,s_T)$, with $s$ connected to its successor $s'$ by the shortest path, with length $d(s, s')$.
As the robot follows a plan, it builds a map $M_{\pi}$ of the environment.
The problem of determining a plan for quality-constrained exploration can then be formulated as
\begin{equation}
    \min_{\pi}\ d(\pi) \quad s.t.\ \mathrm{DesiredCoverage}(M_{\pi}),
    \label{eq:ae}
\end{equation}
where $d(\pi)$ is the total length of the plan.
However, it is typically not possible to evaluate the coverage or feasibility of a complete plan, since at best the environment is partially known.
Therefore, in most cases an exploration plan cannot be obtained offline by solving \cref{eq:ae}.

When performing exploration online, the exploration plan at any given time can be decomposed into three components: the plan already followed up to state $s$, the next state $t$ to be determined, and the unknown optimal plan $\pi^*$ that follows it.
Then, \cref{eq:ae} can be reformulated as the sequential decision problem of determining the optimal next state
\begin{equation}
    t^* = \argmin_{t \in C} \Big(d(s, t) + d(\pi^*)\Big),
    \label{eq:ae-sdp}
\end{equation}
where $C$ are the states that will improve the coverage of the map.
While evaluating $C$ and $d(\pi^*)$ is still not possible without knowing the environment, we now discuss how to address these problems when only the current map and, possibly, local predictions are available.

Like coverage, it is not possible to determine all states $C$ that can improve coverage without knowing the whole environment.
However, it is enough to consider the states $F$ that advance coverage and are nearest to the current state, since any path to $C$ must pass through one of these boundary states.
$F$ can be determined by finding the states that collect at least a minimum gain \cite{duberg-ufoexplorer-2022} or, more commonly, approximated by frontiers \cite{yamauchi-frontier-based-1997}.

Since the states that improve coverage depend on the map and, through it, on the already followed plan, the decision problem suffers from the curse of history and is intractable to solve.
Therefore, determining a good heuristic estimate of $d(\pi^*)$ is one of the fundamental problems of autonomous exploration planning.
Predictive ability of the environment for autonomous exploration should reflect in better heuristic estimates and, consequently, better exploration plans.
However, it has been reported in the literature that existing heuristics do \textit{not} improve with better predictions \cite{ericson-understanding-2021}, indicating there is room for improvement in the planning heuristic.

\subsection{Traditional Exploration Heuristics}

Having identified the sequential decision problem corresponding to quality-constrained exploration, it is now possible to analyze the implicit assumptions of the traditional autonomous exploration heuristics: nearest frontier and information gain.

\textit{Nearest frontier \cite{duberg-ufoexplorer-2022,yamauchi-frontier-based-1997}:}
The exploration plan is obtained through
\begin{equation}
    t^* = \argmin_{t \in F} d(s, t),
    \label{eq:ae-nf}
\end{equation}
implicitly assuming every frontier has an equal-length optimal plan it can follow afterwards.

\textit{Gain maximization \cite{bircher-receding-2016, gonzalez-banos-navigation-2002}:}
The gain $G(s, t)$ is the increase in coverage obtained by following the shortest path from $s$ to $t$.
Its estimate is typically an upper bound \cite{bircher-receding-2016, gonzalez-banos-navigation-2002}, but can be improved using learning-based approaches \cite{ericson-beyond-2024, schmid-fast-2022, shrestha-learned-2019}.
The exploration plan is obtained through 
\begin{equation}
    t^* = \argmax_{t \in F} \Big(\lambda \log G(s, t) - d(s, t)\Big),
    \label{eq:ae-ig}
\end{equation}
where $\lambda$ determines the affinity of gain maximization relative to path cost.
Determining an exploration path by maximizing gain corresponds to assuming that collecting more information on the path to $t$ will lead to a shorter optimal path afterward.

\captionsetup[subfigure]{
  margin=2em,
  %format=hang
}

\begin{figure}[t]
    \centering%
    {\includegraphics[clip,trim=2mm 2mm 2mm 6mm,width=0.9\linewidth]{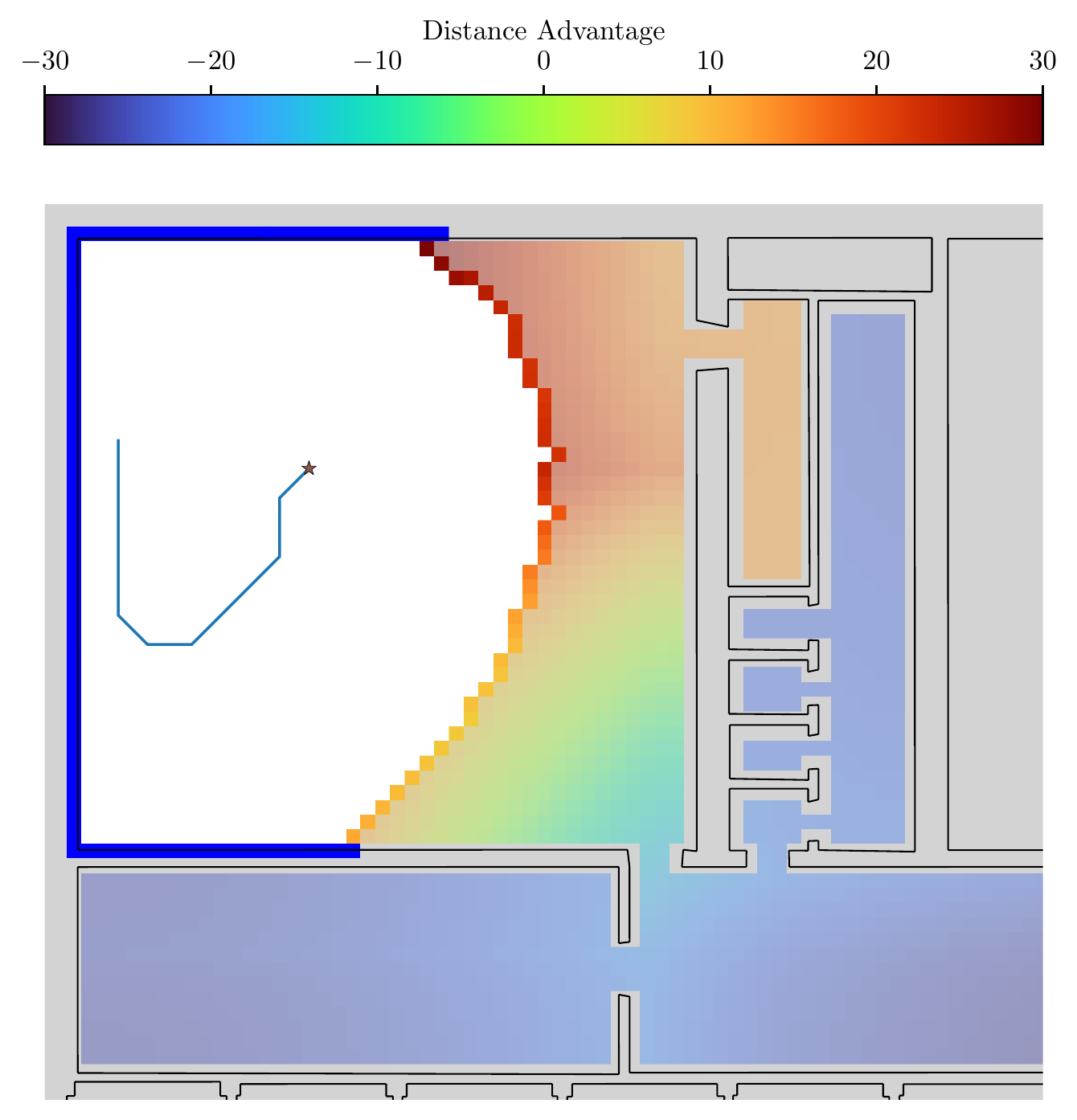}}%
    \caption{Illustration of distance advantage in the beginning of exploration. The robot (star) preferentially explores frontiers (solid coloring) with higher distance advantage. It is heading towards a closed off room because it is nearer that region than it would be from most other places. By contrast, its distance to the corridor is higher than it would be elsewhere, repelling it from that region.}%
    \label{fig:page-one}
\end{figure}

\begin{figure*}
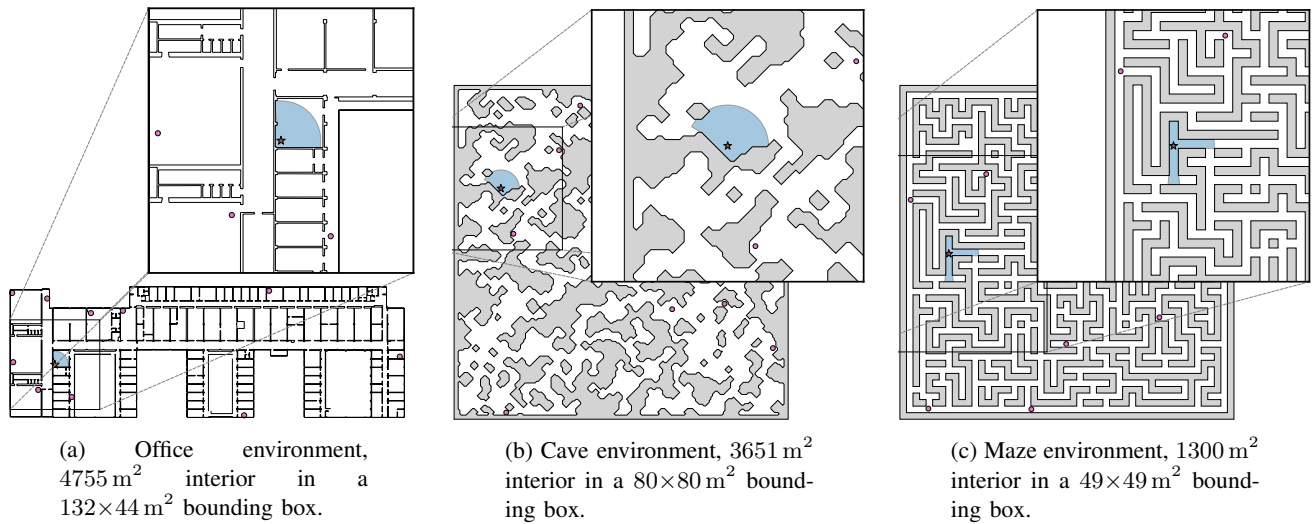

  \center%
  \hfill%
  \begin{subfigure}[t]{0.31\textwidth}%
    \centering%
    \adjustbox{clip,trim=5mm 3mm 3mm 2mm,max width=\textwidth}{\input{office.pgf}}%
    \caption{Office environment, \qty{4755}{\meter^2} interior in a \qtyproduct{132x44}{\meter} bounding box.}
    \label{fig:map-office}
  \end{subfigure}%
  \hfill%
  \begin{subfigure}[t]{0.31\textwidth}%
    \centering%
    \adjustbox{clip,trim=5mm 5mm 2mm 2mm,max width=\textwidth}{\input{cave.pgf}}%
    \caption{Cave environment, \qty{3651}{\meter^2} interior in a \qtyproduct{80x80}{\meter} bounding box.}%
    \label{fig:map-cave}
  \end{subfigure}%
  \hfill%
  \begin{subfigure}[t]{0.31\textwidth}%
    \centering%
    \adjustbox{clip,trim=5mm 5mm 2mm 2mm,max width=\textwidth}{\input{maze.pgf}}%
    \caption{Maze environment, \qty{1300}{\meter^2} interior in a \qtyproduct{49x49}{\meter} bounding box.}%
    \label{fig:map-maze}
  \end{subfigure}%
  \hfill%
  \caption{Data is collected in three diverse environments: a large office from a real-world floor plan with both small cubicles and large lecture halls, a non-rectilinear cave environment with many small pockets, and a labyrinth-like maze with both shallow and deep dead-ends. Pink circles indicate starting locations, the light blue region depicts a sensor scan from the point of view of an example starting location indicated by the brown star polygon. The zoomed region is the same size as a local window for the planner.}
  \label{fig:maps}
\end{figure*}

\section{Distance Advantage}\label{sec:da}

% Information gain as a heuristic for autonomous exploration prioritizes high gain states earlier in exploration.
% In quality-constrained exploration, this corresponds to the implicit assumption that visiting high gain states decreases the length of the optimal plan.
% However, since the total gain to be collected in quality-constrained exploration is fixed, it is unimportant when in the plan gain is collected, as long as it is collected.
% Therefore, it can actually be expected that preferring states by their gain will lead to inefficient plans, since it comes at the cost of undervaluing the length of the plan.

% Nearest frontier can be seen as the limiting case of information gain, in the case that \textit{no preference} is given to gain, and therefore would be expected to lead to more efficient plans.
% However, it corresponds to a completely uninformative estimate of the advantage of visiting a frontier state, since the implicit assumption is that the length of the optimal path after each frontier is the same for all of them.
% From the map built by exploring the environment and from additional information that might be available, it should be possible to get a better estimate of which frontiers are better to visit.

In order to design a quality-constrained exploration planner, a suitable heuristic has to be found.
% It has already been observed in the literature that unnecessary backtracking is a main source of inefficiency in quality-constrained exploration, for both nearest frontier and information gain \cite{ericson-understanding-2021, obwald-speeding-up-2016}.
It has already been pointed out in \cite{obwald-speeding-up-2016} that unnecessary backtracking happens when a frontier state is not explored by the \textit{last} time that the optimal plan comes close to it.
Therefore, a suitable heuristic for quality-constrained exploration is one that minimizes backtracking by determining which frontiers are unlikely to be revisited.

An indication that a frontier might not be revisited is that it is isolated, i.e., it has high average distance to other reachable states $R$.
Therefore, we propose to determine an exploration plan through
\begin{equation}
    t^* = \argmax_{t \in F} \left(\frac{1}{|R|}\sum_{s'\in R} d(t, s') - d(s, t)\right),
    \label{eq:ae-da}
\end{equation}
where $|R|$ is the number of reachable states.
This heuristic prefers to visit frontiers that are near the robot, but on average isolated from reachable space, \ie, frontiers that the robot has a \textit{distance advantage} to visit.

The reachable states are determined using the map and, if available, map predictions, which we show lead to an improvement in the estimate.
Determining an exploration path through \cref{eq:ae-da} corresponds to assuming that the cost of the optimal path after the frontier is lower if the frontier is more isolated, since leaving it behind would cause backtracking.
In \cref{fig:page-one}, an example exploration state in an occupancy grid environment is shown, with unexplored cells colored according to their distance advantage and frontier cells highlighted with solid coloring.

Computing the distance advantage requires determining the shortest path distance from every frontier to every reachable state.
In a graph-like environment, computing shortest path distances requires as many single-source shortest path problems as there are frontiers, and the computational cost for each scales with the size of the graph, i.e., the number of reachable states.
Therefore, in order to keep the computational cost bounded regardless of the environment size, only frontiers and reachable states that are in a local window centered around the current state are considered.
If the environment is a weighted graph, $d(t, s')$ is determined using Dijkstra's algorithm, while breadth-first search is used for unweighted graphs.

Frontiers that are outside the local window are not considered unless the local window has been fully explored, in which case the fallback planning chooses the nearest frontier outside the local window.

\section{Experimental Setup} \label{sec:experimental-setup}

In \cref{sec:results,sec:predictions}, experiments are conducted to evaluate distance advantage, nearest frontier, and information gain with respect to their quality-constrained exploration performance, and their sensitivity to predictions.
The experiments are conducted in simulation in order to fairly compare the planning objectives.
This section describes the simulation environment and other relevant implementation details.

\subsection{Sensor \& Mapping}

The robot is simulated as a point-like sensor, with holonomic motion capabilities.
The simulated sensor is a \qty{360}{\degree} laser range sensor, with 720 evenly spaced rays and a maximum range of \qty{4.5}{\meter{}}.
As the robot moves and collects sensor scans, these are accumulated into an occupancy grid map that discretizes the environment into \qtyproduct{25 x 25}{\centi\meter} cells and marks them as either free, occupied, or unknown.
The scans are accumulated conservatively, such that 8-connected paths through free space in the map are guaranteed to be collision-free.

% The robot is simulated as a point-like holonomic sensor. 
% If it is important to consider the footprint of the robot, this can be accomplished by inflating occupied space with the robot's footprint.

% The simulated sensor is a \qty{360}{\degree{}} laser range sensor, with 720 evenly spaced rays and a maximum range of \qty{4.5}{\meter{}}.
% The sensor is simulated by casting rays from the robot's position, with the first ray's direction being randomly chosen at the start of the simulation.
% For each ray, its detected range is the one at which it intersects an opaque boundary or the maximum range, if it does not intersect.
% For transparent boundaries, like windows, the ray does not terminate when crossing the boundary, but the range is also recorded.
% Thus, a ray that crosses a transparent boundary has multiple detected ranges.
% In \cref{fig:sensor}, an example simulated sensor scan is shown.

% The sensor scans are accumulated into an occupancy grid map, which discretizes the environment into \qtyproduct{25 x 25}{\centi\meter} cells that are classified as either free, occupied or unknown.
% The intricacies of occupancy grid mapping are discussed in \cref{ap:ogm}.
% In \cref{fig:ogm}, the occupancy grid map resulting from the sensor scan of \cref{fig:sensor} is shown.

\subsection{Localization \& Path Execution}

When the next state is chosen by the planner, a shortest 8-connected path is found through exhaustive search in the map using Dijkstra's algorithm.
Since there are often many such shortest paths, the one which keeps the most distance from walls is chosen.
As the path is executed, at each step in the map, a new sensor scan is obtained.
The path terminates when the goal state is reached or when the sensor scan causes a map cell that was unknown to be marked as occupied or free.

Since the path is executed deterministically, the robot is perfectly localized with respect to the starting state.

\subsection{Predictions}

A map predictor, like the ones from \cite{ericson-beyond-2024, luperto-mapping-2023, shrestha-learned-2019}, is assumed to be available and capable of providing map completions in a local window centered around the current state.
The map predictor extends the map with the real environment and the local window size is \qtyproduct{30x30}{\meter}.

\section{Quality-Constrained Evaluation}\label{sec:results}

For the purposes of autonomous exploration, environments can be characterized by their connectivity.
One end of the spectrum is a perfectly tree-like environment, \eg, a maze without loops.
In this case, it is practically irrelevant which frontier is chosen, as long as it is explored to completion, since the robot must always return to the branching point to pursue the next frontier.
On the other end of the spectrum is a fully connected environment, where the unexplored space beyond every frontier eventually connects.
Real environments are somewhere in the middle of this spectrum, with some branches interconnected and some branches dead ends, \eg, an office environment or a cave system.
% NF is near optimal as the decisions at each branching point do not matter; the robot must return to that branching point once it has completed whatever branch it took. On the other end is the fully connected environment, where every branch is connected to every other, so the choice does not matter. Real environments are somewhere in the middle of this spectrum, with some branches interconnected and some branches dead ends.

The three heuristics are evaluated in three environments with different characteristics, presented in \cref{fig:maps}: a large office, with a wide variety of rooms-within-rooms, looping corridors, and wide-open spaces; a cave-like environment with high global connectivity, but low local connectivity; and a maze with a relatively small amount of loops, mostly consisting of dead-end branches of varying depth.
The distance at completion for all heuristics in each environment is shown in \cref{tab:results}.
Since nearest frontier is a special case of information gain (IG), with no preference for information, it is used as a reference.

\begin{table}[tb]
\centering
\caption{Distance at completion for each method in each environment of \cref{fig:maps}. Data collected across 10 runs for each method/environment pair from different starting locations.
The difference to nearest frontier is computed per starting location.
}
\label{tab:results}
\def\NF{Nearest Frontier}
\begin{tabular}{@{}l S[table-format=4.1(2.1)] S[table-format=4.1(3.1)] S[table-format=4.1(2.1)]@{}}
\toprule
 & \multicolumn{3}{c}{Distance at Completion $d_T$ (\si{\meter})} \\
\cmidrule(lr){2-4}
{Method} & {Office} & {Cave} & {Maze} \\
\midrule
Nearest Frontier & 1892.5(50.5) & 1854.3(106.1) & 1377.8(36.0) \\
\midrule
Dist. Adv.       & 1578.1(43.9) &  1652.1(93.4) & 1249.5(27.5) \\
$\Delta$\NF{}    & -303.6(35.0) &  -194.8(68.8) & -128.3(45.3) \\
\midrule
IG Max. & 2330.3(71.8) & 2313.2(133.0) & 1514.1(40.9) \\
$\Delta$\NF{}    &  437.8(89.8) &  407.5(102.1) &  136.3(60.9) \\
\bottomrule
\end{tabular}
\end{table}

In the office environment, which has the most complex connectivity of the three, distance advantage obtains an improvement of \pct{16} over nearest frontier, while information gain is \pct{23} worse.
As the connectivity of the environments simplifies, the margin for improvement or degradation over nearest frontier decreases, since the simpler connectivity leads to there being more opportunities for sub-optimal policies to correct mistakes at a low cost.
These results show that distance advantage consistently obtains shorter paths, with the difference being more noticeable in environments with more complex connectivity.
Information gain is consistently outperformed by nearest frontier and, by extension, distance advantage.

\subsection{Greediness}

\begin{figure}[tb]
  \centering%
  \begin{adjustbox}{clip,trim=2mm 2mm 2mm 2mm,width=\linewidth}%
    \input{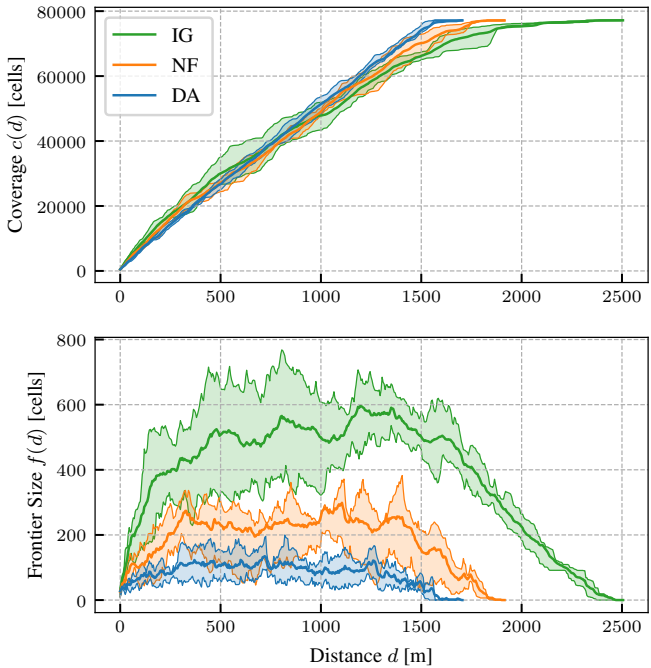}%
  \end{adjustbox}%
  \caption{Comparison of coverage $c(d)$ and total frontier size $f(d)$ as functions of distance traveled $d$. The shaded areas indicates an \pct{80} confidence interval, the solid line indicates the mean. Data collected across $10$ runs for each method from different starting locations in the office environment.}
  \label{fig:docod}
\end{figure}

Information gain has previously been shown to lead to greedy behaviors, due to sacrificing long-term exploration performance for early short-term gains \cite{ericson-understanding-2021, obwald-speeding-up-2016}. 
In order to evaluate how greediness affects performance, the coverage $c(d)$ and frontier size $f(d)$ as functions of the distance traveled $d$ were considered, as is shown in \cref{fig:docod}.

There are nearly no differences in the coverage rate early on, indicating that the information gain maximization strategy is not successful in doing so.
However, there is a large difference in the number of the frontiers, with nearest frontier quickly growing to double the amount of distance advantage, and gain maximization more than doubling nearest frontier.
In accordance with \cite{ericson-understanding-2021}, these outstanding frontiers represent a kind of debt to be paid, in the form of travel distance at the end of the run.
This outstanding frontier debt explains why the coverage rate for information gain, and to a lesser extent nearest frontier, decreases as exploration progresses and ultimately leads to a longer path.
By contrast, the frontier size for distance advantage remains approximately constant during most of the exploration run, allowing it to keep an almost constant coverage rate until the end of exploration.

%can be observed, it is not significant, and the range of each method overlaps  lies within the distribution of runs.

\subsection{Gain Maximization Affinity}

When first introducing gain maximization for autonomous exploration, \cite{gonzalez-banos-navigation-2002} already highlighted that lower affinity $\lambda$ led to a more meticulous covering of the environment.
We evaluate the effect of gain maximization affinity on path length with two gain estimators: naive, which assumes unknown space is non-occluding, and true, which has access to the true gain.
The resulting completion distances are shown in \Cref{fig:ig-lambda}, clearly showing that higher affinity leads to longer paths and that the effect is worsened with more accurate estimates.
% The gain maximization affinity $\lambda$ in \cref{eq:ae-ig} regulates the relative preference of gain over frontier proximity.

An interesting effect that, to the best of the authors' knowledge, has not been reported is that \textit{negative} affinity leads to shorter paths than nearest frontier.
The fact that preference for lower gain improves performance can be understood through the lens of isolation, since a low predicted gain amounts to predicting that a frontier region is soon to terminate.
In that way, the preference for low gain is an \textit{ad hoc} heuristic for preferring shallow frontiers, that are unlikely to be revisited by the optimal path since they do not continue deeper into unknown space.

\section{Sensitivity to Predictions}\label{sec:predictions}

The previous experiments all consider perfect predictions, given by an oracle that knows the true environment.
However, in a real exploration scenario, these predictions would come either from prior knowledge of the environment, e.g., a floor plan, or from learning-based models \cite{ericson-beyond-2024, luperto-mapping-2023, shrestha-learned-2019}.
In those cases, the predictions will not perfectly reflect the environment, and it is important to assess the sensitivity of each method to the predictions, with respect to both the amount of predicted information and its accuracy.
Nearest frontier is completely insensitive to predictions and is presented as a baseline.

\subsection{Prediction Range}

The amount of information made available through the predictor could have a large influence in the exploration plan.
While \cite{ericson-beyond-2024} showed that it is possible to predict \qtyproduct{30x30}{\meter} windows from the occupancy map built by the robot in office environments, this might not be possible in more complex environments or too computationally demanding in some situations.
Therefore, we examine the effect of the prediction range beyond the frontier, parameterized by the maximum straight line distance $c_p$ in number of cells.

\begin{figure}[tb]
  \centering%
  \begin{adjustbox}{clip,trim=2mm 3mm 1mm 2mm,width=\linewidth}%
    \input{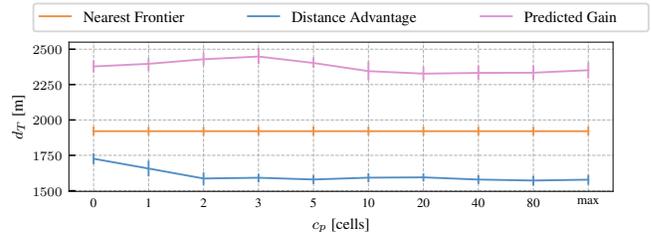}%
  \end{adjustbox}%
  \caption{The effect of prediction range $c_p$ on completion distance $d_T$ in the office environment. Data collected across 10 runs from different starting locations, for each method and prediction range. Error bars represent one standard deviation.}%
  \label{fig:c-p}
\end{figure}

\Cref{fig:c-p} illustrates how different settings of $c_p$ affect performance for all methods in the office environment.
It can be seen that the ranking of the methods does not change in the absence of predictions ($c_p=0$), and distance advantage continues to outperform nearest frontier.
Most of the gain that distance advantage gets from the predictions is already realized at $c_p=2$, which corresponds to a prediction range of only $\sim\qty{50}{\centi\meter}$ beyond the frontier.
In agreement with the results reported by \cite{ericson-understanding-2021}, information gain fails to take advantage of predictions.

% Since we rely on predictions to improve performance, it is natural to ask to what extent these predictions are critical to our method. To examine this, we limit how far beyond the frontier the robot can predict, parameterized by the maximum straight line distance $c_p$. 
% . As expected, increasing $c_p$ consistently improves performance. Specifically, we see an $\sim \qty{0}{\meter} (\pct{8.5})$ improvement when using full predictions ($c_p \to \infty$) compared to having no predictions ($c_p = 0$).

% Interestingly, even for $c_p = 0$, the DA method outperforms NF by $\sim \qty{0}{\meter}$ (\pct{8}). This suggests that not only are predictions beneficial to DA, but that DA provides advantages even in zero-knowledge autonomous exploration settings.
% NF -> DA c_p 0 ~ 8%
% DA c_p 0 -> None ~ 8.5%

% NF               1857.682658
% DA $c_p=0$       1709.364664
% DA $c_p=5$       1669.544846
% DA $c_p=10$      1618.653706
% DA $c_p=20$      1618.980930
% DA $c_p=40$      1596.615791
% DA $c_p=None$    1563.683123

\begin{figure}[!tb]
  \center%
  \fbox{\adjustbox{clip,trim=48mm 64mm 6mm 19mm,max width=\linewidth}{\input{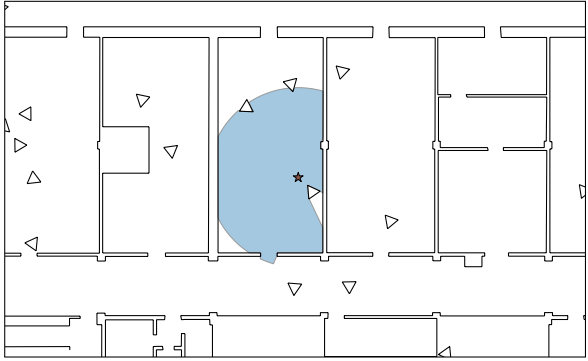}}}%
  \caption{An example of clutter for the office environment. The clutter consists of polygons, roughly \qty{1}{\meter}, randomly placed in the reachable environment. A section of the environment does not get disconnected from the rest due to the clutter closing a passage, like the clutter of actual human environments.}
  \label{fig:map-office-clutter}
\end{figure}

\subsection{Prediction Accuracy}

A common mode of failure for predictions is non-structural elements of the environment such as furniture, human occupants, other robots, etc., since those are likely to move or be moved around the environment.
In the office environment, the floor plan is fixed and predictions of it might be accessible, but elements like desks, chairs, etc., can be unpredictable.
A different source of prior knowledge could also be an old map, in which case the mismatch could go the other way; the clutter in the map might no longer reflect the true state of the environment.
To assess the impact of this kind of mismatch between predictions and the actual environment, clutter was generated for the office environment in~\cref{fig:map-office} by randomly sampling triangles, illustrated in~\cref{fig:map-office-clutter}. The same set of triangles is used for every method, but different sets are used between starting locations, predictions and the true environment.
The results of this evaluation are presented in \cref{tab:noise-results}.

\begin{table}[tb]
\centering
\caption{Distance at completion when there is mismatch between the predictions and the environment, due to clutter. The environment and prediction clutter are independently sampled. Data collected across 10 runs for each method/environment/prediction tuple from different starting locations.}
\label{tab:noise-results}
\def\NF{Nearest Frontier}
\begin{tabular}{@{}ccc S[table-format=4.1(2.1)]@{}}
\toprule
{Method} & {Environment} & {Prediction} & {$d_T$ (\si{\meter})}\\
\midrule
     & Clean & Clean & 1892.5(50.5) \\
 \NF & Noise & Clean & 2041.7(57.4) \\
     & Noise & Noise & 2054.0(32.1) \\
\midrule
            & Clean & Clean & 1578.1(43.9) \\
 Dist. Adv. & Noise & Clean & 1782.3(52.1) \\
            & Noise & Noise & 1812.0(14.8) \\
\midrule
            & Clean & Clean & 2351.6(85.2) \\
 IG Max.    & Noise & Clean & 2567.8(76.8) \\
            & Noise & Noise & 2563.0(85.6) \\
\bottomrule
% & \multicolumn{2}{c}{Distance at Completion $d_T$ (\si{\meter})} \\
%\cmidrule(lr){2-3}
%{Method} & {Clean Env./Clutter Pred.} & {Clutter Env./Clutter Pred.} \\
%\midrule
%\NF{} & 1892.5(50.5) & 1854.3(106.1) \\
%\midrule
%Dist. Adv.       & 1578.1(43.9) &  1652.1(93.4) \\
%%$\Delta$\NF{}    & -303.6(35.0) &  -194.8(68.8) \\
%\midrule
%Information Gain & 2330.3(71.8) & 2313.2(133.0) \\
%%$\Delta$\NF{}    &  437.8(89.8) &  407.5(102.1) \\
%\bottomrule
\end{tabular}
\end{table}

% DA & CT-CM & 1578.1(43.9) \\
% IG & CT-CM & 2351.6(85.2) \\

% NF & NT-CM & 2041.7(57.4) \\
% DA & NT-CM & 1782.3(52.1) \\
% IG & NT-CM & 2567.8(76.8) \\

% NF & NT-NM & 2054.9(32.1) \\
% DA & NT-NM & 1812.2(14.8) \\
% IG & NT-NM & 2563.2(85.6) \\

Clutter is a source of occlusion, so when the environment is cluttered it is harder to observe the environment from far away, causing all methods to produce longer paths.
While nearest frontier produces ${\sim\qty{150}{\meter}}$ longer paths, distance advantage and information gain degrade by ${\sim\qty{200}{\meter}}$.
This suggests that the $\qty{150}{\meter}$ increase can be explained due to the cluttering of the environment making observing it harder, with the inaccuracy in predictions accounting for the remaining $\qty{50}{\meter}$.
Importantly, the way in which the predictions are wrong, \ie, whether they do not contain clutter or they contain mismatched clutter, does not have a significant influence on the performance.

%door closedness 0, 25, 50, 75, 100 percent: optimistic vs pessimistic about reachability

% four-way
\iffalse
+------------+---------+-----------+
| act \ pred | clutter |   clean   |
+------------+---------+-----------+
|  clutter   |    -    | realistic |
|   clean    | old map |   basic   |
+------------+---------+-----------+
\fi

\section{Limitations}\label{sec:limitations}

The main focus of this work is to highlight how the current state-of-the-art autonomous exploration planning methods ultimately do not optimize the correct objectives.
Although the proposed objective is well-motivated both by intuition and by empirical results, a more thorough theoretical analysis has the potential to yield even better formulations.
Future work should attempt to characterize the fundamental limits of the proposed objective and improve upon it.
Another aspect that warrants additional work is the computational scalability of estimating the optimization objective.
Since the proposed objective requires solving a multi-source shortest path problem, the computational time scales with the size of the explored map.
This issue was addressed by limiting the computation to a fixed-size local map, but future work should investigate whether it is possible to improve upon this solution.

The experimental results are limited to the minimal case, with no uncertainty and perfect mapping.
%To further confirm the findings presented in this paper, additional experiments should be performed with reduced simplifications.
%This includes considering simulations with motion and observation uncertainty, thereby requiring SLAM, considering predictions produced by data-driven models, and real-world experiments.

\section{Conclusion}\label{sec:conclusion}

In this work, we highlight the important differences between budget- and quality-constrained exploration, and address the inconsistencies observed in the autonomous exploration community.
We investigate why traditional heuristics, such as information gain and nearest frontier, perform poorly in the quality-constrained paradigm, and propose a new heuristic for quality-constrained exploration.
Since quality-constrained exploration is defined to be completed when the map is of sufficient quality, total gain is fixed; maximizing information gain of individual frontiers is therefore ultimately irrelevant as all the gain will be collected eventually.
Therefore, the central problem in quality-constrained exploration planning is to determine the correct order in which to explore frontiers, such that the length of unnecessary detours is minimized.

We propose a heuristic, named \textit{distance advantage}, that attempts to identify which frontiers have higher opportunity cost if missed, i.e., those more likely to require a detour later in exploration if they are not explored now, by estimating their average distance to other states.
This heuristic is compared to nearest frontier and information gain, and it is shown to consistently explore the environments with shorter paths.
Perhaps most importantly, among the evaluated heuristics, distance advantage is the only one to show improvements in performance as access to predictions improves, and is able to handle imperfect predictions.
We believe these results clearly show the different nature of budget- and quality-constrained exploration, and indicate that further work should be done in understanding the correct objective for quality-constrained exploration, and exploring the proposed objective.

% We discussed how in the quality-constrained exploration paradigm, greedy objectives like information gain are no longer well suited for the task of exploration planning and 
% In this work, the problem of choosing an adequate planning objective for autonomous exploration was investigated.
% Information gain as a path-planning objective was shown to be consistently detrimental to autonomous exploration performance.
% Although better than information gain, nearest-frontier exploration was also shown to be susceptible to improvement.
% A new objective, based on considering not only the distances to frontiers, but also the distances \emph{between} frontiers, was proposed.
% This objective was shown to outperform both information gain and nearest frontier.
    
% back matter

\balance
\bibliographystyle{IEEEtrandiy}
\bibliography{main}

\begin{thebibliography}{10}
\providecommand{\url}[1]{#1}
\csname url@rmstyle\endcsname
\providecommand{\newblock}{\relax}
\providecommand{\bibinfo}[2]{#2}
\providecommand\BIBentrySTDinterwordspacing{\spaceskip=0pt\relax}
\providecommand\BIBentryALTinterwordstretchfactor{4}
\providecommand\BIBentryALTinterwordspacing{\spaceskip=\fontdimen2\font plus
\BIBentryALTinterwordstretchfactor\fontdimen3\font minus
  \fontdimen4\font\relax}
\providecommand\BIBforeignlanguage[2]{{%
\expandafter\ifx\csname l@#1\endcsname\relax
\typeout{** WARNING: IEEEtran.bst: No hyphenation pattern has been}%
\typeout{** loaded for the language `#1'. Using the pattern for}%
\typeout{** the default language instead.}%
\else
\language=\csname l@#1\endcsname
\fi
#2}}

\bibitem{calisi-autonomous-2007}
D.~Calisi, A.~Farinelli, L.~Iocchi, and D.~Nardi, ``Autonomous exploration for
  search and rescue robots,'' \emph{Tran. Built Environment}, 2007.

\bibitem{colas-3d-2013}
F.~Colas, S.~Mahesh, F.~Pomerleau, M.~Liu, and R.~Siegwart, ``3d path planning
  and execution for search and rescue ground robots,'' in \emph{Proc. Int.
  Conf. Intell. Robot. Syst.}\hskip 1em plus 0.5em minus 0.4em\relax IEEE,
  2013.

\bibitem{omari-visual-2014}
S.~Omari, P.~Gohl, M.~Burri, M.~Achtelik, and R.~Siegwart, ``Visual industrial
  inspection using aerial robots,'' in \emph{Proc. Int. Conf. Appl. Robot.
  Power Ind.}\hskip 1em plus 0.5em minus 0.4em\relax IEEE, 2014.

\bibitem{ericson-understanding-2021}
L.~Ericson, D.~Duberg, and P.~Jensfelt, ``Understanding greediness in
  map-predictive exploration planning,'' in \emph{Proc. Eur. Conf. Mob.
  Robot.}\hskip 1em plus 0.5em minus 0.4em\relax IEEE, 2021, pp. 1--7.

\bibitem{luperto-estimating-2024}
M.~Luperto, M.~M. Ferrara, G.~Boracchi, and F.~Amigoni, ``Estimating map
  completeness in robot exploration,'' 2024.

\bibitem{obwald-speeding-up-2016}
S.~Obwald, M.~Bennewitz, W.~Burgard, and C.~Stachniss, ``Speeding-up robot
  exploration by exploiting background information,'' \emph{Robot. Autom.
  Letters}, 2016.

\bibitem{ericson-beyond-2024}
L.~Ericson and P.~Jensfelt, ``Beyond the frontier: Predicting unseen walls from
  occupancy grids by learning from floor plans,'' \emph{Robot. Autom. Letters},
  2024.

\bibitem{yamauchi-frontier-based-1997}
B.~Yamauchi, ``A frontier-based approach for autonomous exploration,'' in
  \emph{Proc. Int. Symp. Comput. Intell. Robot. Autom.}\hskip 1em plus 0.5em
  minus 0.4em\relax IEEE, 1997, pp. 146--151.

\bibitem{gonzalez-banos-navigation-2002}
H.~H. González-Baños and J.-C. Latombe, ``Navigation strategies for exploring
  indoor environments,'' \emph{Int. Journal Robot. Research}, 2002.

\bibitem{bajcsy-active-1988}
R.~Bajcsy, ``Active perception,'' \emph{Proc. IEEE}, 1988.

\bibitem{whaite-autonomous-1997}
P.~Whaite and F.~P. Ferrie, ``Autonomous exploration: Driven by uncertainty,''
  \emph{Tran. Pat. Anal. Machine Intell.}, vol.~19, no.~3, pp. 193--205, 1997.

\bibitem{connolly-determination-1985}
C.~Connolly, ``The determination of next best views,'' in \emph{Proc. Int.
  Conf. Robot. Autom.}\hskip 1em plus 0.5em minus 0.4em\relax IEEE, 1985.

\bibitem{maver-occlusions-1993}
J.~Maver and R.~Bajcsy, ``Occlusions as a guide for planning the next view,''
  \emph{Tran. Pattern Analysis Machine Intell.}, 1993.

\bibitem{pito-sensor-based-1996}
R.~Pito, ``A sensor-based solution to the "next best view" problem,'' in
  \emph{Proc. Int. Conf. Pattern Recognit.}\hskip 1em plus 0.5em minus
  0.4em\relax IEEE, 1996.

\bibitem{bircher-receding-2016}
A.~Bircher, M.~Kamel, K.~Alexis, H.~Oleynikova, and R.~Siegwart, ``Receding
  horizon "next-best-view" planner for 3d exploration,'' in \emph{Proc. Int.
  Conf. Robot. Autom.}\hskip 1em plus 0.5em minus 0.4em\relax IEEE, 2016.

\bibitem{selin-efficient-2019}
M.~Selin, M.~Tiger, D.~Duberg, F.~Heintz, and P.~Jensfelt, ``Efficient
  autonomous exploration planning of large-scale {3-D} environments,''
  \emph{Robot. Autom. Letters}, 2019.

\bibitem{kulich-2011-distance}
M.~Kulich, J.~Faigl, and L.~P{\v{r}}eu{\v{c}}il, ``On distance utility in the
  exploration task,'' in \emph{Proc. Int. Conf. Robot. Autom.}\hskip 1em plus
  0.5em minus 0.4em\relax IEEE, 2011.

\bibitem{zhou-fuel-2021}
B.~Zhou, Y.~Zhang, X.~Chen, and S.~Shen, ``{FUEL}: Fast {UAV} exploration using
  incremental frontier structure and hierarchical planning,'' \emph{Robot.
  Autom. Letters}, 2021.

\bibitem{schmid-efficient-2020}
L.~Schmid, M.~Pantic, \emph{et~al.}, ``An efficient sampling-based method for
  online informative path planning in unknown environments,'' \emph{Robot.
  Autom. Letters}, 2020.

\bibitem{duberg-ufoexplorer-2022}
D.~Duberg and P.~Jensfelt, ``{UFOExplorer}: Fast and scalable sampling-based
  exploration with a graph-based planning structure,'' \emph{Robot. Autom.
  Letters}, 2022.

\bibitem{jiajie-echo-2023}
J.~Yu, H.~Shen, J.~Xu, and T.~Zhang, ``{ECHO}: An efficient heuristic viewpoint
  determination method on frontier-based autonomous exploration for
  quadrotors,'' \emph{Robot. Autom. Letters}, 2023.

\bibitem{lindqvist-explorationrrt-2021}
B.~Lindqvist, A.-A. Agha-Mohammadi, and G.~Nikolakopoulos, ``Exploration-rrt: A
  multi-objective path planning and exploration framework for unknown and
  unstructured environments,'' in \emph{Proc. Int. Conf. Intell. Robot.
  Syst.}\hskip 1em plus 0.5em minus 0.4em\relax IEEE, 2021.

\bibitem{witting-history-2018}
C.~Witting, M.~Fehr, R.~B{\"a}hnemann, H.~Oleynikova, and R.~Siegwart,
  ``History-aware autonomous exploration in confined environments using mavs,''
  in \emph{Proc. Int. Conf. Intell. Robot. Syst.}\hskip 1em plus 0.5em minus
  0.4em\relax IEEE, 2018.

\bibitem{deng-robotic-2020}
D.~Deng, R.~Duan, J.~Liu, K.~Sheng, and K.~Shimada, ``Robotic exploration of
  unknown {2D} environment using a frontier-based automatic-differentiable
  information gain measure,'' in \emph{Proc. Int. Conf. Adv. Intell.
  Mechatron.}\hskip 1em plus 0.5em minus 0.4em\relax IEEE, 2020.

\bibitem{schmid-fast-2022}
L.~Schmid, C.~Ni, Y.~Zhong, R.~Siegwart, and O.~Andersson, ``Fast and
  compute-efficient sampling-based local exploration planning via distribution
  learning,'' \emph{Robot. and Autom. Letters}, 2022.

\bibitem{shrestha-learned-2019}
R.~Shrestha, F.-P. Tian, W.~Feng, P.~Tan, and R.~Vaughan, ``Learned map
  prediction for enhanced mobile robot exploration,'' in \emph{Proc. Int. Conf.
  Robot. Autom.}\hskip 1em plus 0.5em minus 0.4em\relax IEEE, 2019.

\bibitem{tao-seer-2023}
Y.~Tao, Y.~Wu, \emph{et~al.}, ``{SEER}: Safe efficient exploration for aerial
  robots using learning to predict information gain,'' in \emph{Proc. Int.
  Conf. Robot. Autom.}\hskip 1em plus 0.5em minus 0.4em\relax IEEE, 2023.

\bibitem{li-searching-2012}
A.~Q. Li, F.~Amigoni, and N.~Basilico, ``Searching for optimal off-line
  exploration paths in grid environments for a robot with limited visibility,''
  vol.~26, no.~1, pp. 2060--2066, 2012.

\bibitem{stachniss-information-2005}
C.~Stachniss, G.~Grisetti, and W.~Burgard, ``Information gain-based exploration
  using {Rao}-{Blackwellized} particle filters,'' in \emph{Proc. Robot.: Sci.
  Syst.}\hskip 1em plus 0.5em minus 0.4em\relax The MIT Press, 2005.

\bibitem{yichen-exploration-2022}
Y.~Zhang, B.~Zhou, L.~Wang, and S.~Shen, ``Exploration with global consistency
  using real-time re-integration and active loop closure,'' in \emph{Proc. Int.
  Conf. Robot. Autom.}\hskip 1em plus 0.5em minus 0.4em\relax IEEE, 2022.

\bibitem{luperto-mapping-2023}
M.~Luperto, F.~Amadelli, M.~Di~Berardino, and F.~Amigoni, ``Mapping beyond what
  you can see: Predicting the layout of rooms behind closed doors,''
  \emph{Robot. Autonomous Syst.}, 2023.

\end{thebibliography}

% \appendix

% \section{Occupancy Grid Maps}\label{ap:ogm}
% \zemark{say that a good orientation is chosen to remove problems with occupancy grids and axis alignment.}

% In order to ensure that plans synthesized in the map produced by the robot are feasible, i.e., they cannot lead to collisions, the sensor scans must be converted to occupancy grid scans conservatively.
% \zemark{things about max range, neighbours, etc.}
% \zemark{mention axis recovering?}
% In \cref{fig:sensor}, an example sensor scan is shown, alongside the conservatively synthesized occupancy grid scan.
% The occupancy grid map is built by accumulating the occupancy grid scans from each sensor scan.

\end{document}